# COMBINED STATISTICAL AND MODEL BASED TEXTURE FEATURES FOR IMPROVED IMAGE CLASSIFICATION

Omar Al-Kadi*

*University of Sussex, Brighton, United Kingdom (o.al-kadi@sussex.ac.uk)

## Abstract

This paper aims to improve the accuracy of texture classification based on extracting texture features using five different texture methods and classifying the patterns using a naïve Bayesian classifier. Three statistical-based and two model-based methods are used to extract texture features from eight different texture images, then their accuracy is ranked after using each method individually and in pairs. The accuracy improved up to 97.01% when model based – Gaussian Markov random field (GMRF) and fractional Brownian motion (fBm) – were used together for classification as compared to the highest achieved using each of the five different methods alone; and proved to be better in classifying as compared to statistical methods. Also, using GMRF with statistical based methods, such as Gray level co-occurrence (GLCM) and run-length (RLM) matrices, improved the overall accuracy to 96.94% and 96.55%; respectively.

## 1. Introduction

IMAGE texture represents the appearance of the surface and how its elements are distributed. It is considered an important concept in machine vision, in a sense it assists in predicting the feeling of the surface (e.g. smoothness, coarseness …etc) from image.

Various texture analysis approaches tend to represent views of the examined textures form different perspectives, and due to multi-dimensionality of perceived texture, there is not an individual method that can be sufficient for all textures [1]. Therefore, this work is mainly concerned with texture classification accuracy improvement using textures features derived from model and statistical based methods.

In the model-based approach, a set of parameters which are driven from the variation of pixel elements of texture are used to define an image model. The two models methods used in this work are Gaussian Markov random field (GMRF) and fractional Brownian motion (fBm), where the former sets the conditional probability of the intensity of a certain pixel depending on the values of the neighbouring pixels while the latter exploits the self-similarity of texture at varying scales. For statistical-based methods, first and second order statistics is derived after analyzing the spatial distributions of pixel grey level values. Gray level co-occurrence, run-length and autocovariance function methods were selected for feature extraction.

The obtained texture features by different methods are used individually and in combination with each other for classification. A supervised learning approach was adopted for training and testing the extracted features from samples of image segments obtained from each image class using a naïve Bayesian classifier.

## 2. Methodology

### 2.1 Data set preparation

Different type of texture images ranging from fine to coarse for the purpose of texture classification were used in this paper as shown in Fig.1 . Eight different texture images having size of 256x256 with 8-bit grey levels were selected from the Brodatz album [2]. Each image which defines a separate class was divided into size of 32 x 32 image segments with 50% overlapping. Nearly one third of the images segments (64 samples) referring to each class was used for training, while the rest (192 samples) was used for testing the classifier.

### 2.2 Texture features extraction

Five different methods – 2 model and 3 statistical based – were used to extract different texture features from 2048 image segments samples referring to 8 different texture classes, as follows:-

*2.2.1 Model-based features methods*

*Random fields*

Based upon the Markovian property, which is simply the dependence of each pixel in the image on its neighbors only, a Gaussian Markov random field model (GMRF) for third order Markov neighbors was used [3]. The 7 GMRF parameters are estimated using least square error estimation method.

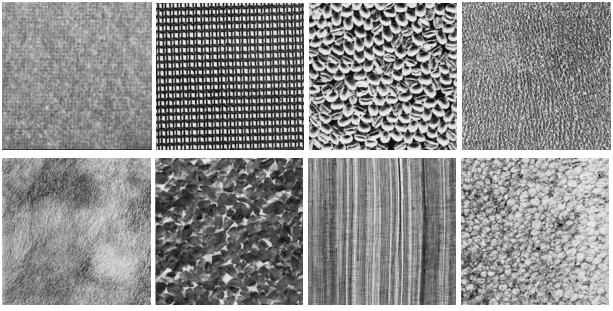

Fig. 1. Eight different Brodatz texture images showing from up to bottom and from left to right: herringbone cloth (D16), canvas (D20), coffee beans (D74), calf leather (D24), fur (D93), quartz (D98), cheese cloth (D106) and plastic bubbles (D112).

The GMRF model is defined by the following formula:

$$p(I_{ij} | I_{kl}, (k,l) \in N_{ij}) = \frac{1}{\sqrt{2\pi\sigma^2}} \exp\left\{\frac{\left(I_{ij} - \sum_{l=1}^{n} \alpha_l s_{kl;l}\right)^2}{2\sigma^2}\right\} \quad (1)$$

Where the right hand side of (1) represents the probability of a pixel (i,j) having a specific grey value $I_{ij}$, given the values of its neighbors, $n$ is the total number of pixels in the neighborhood $N_{ij}$ of pixel $I_{ij}$, which influence its value, $\alpha_l$ is the parameter with which a neighbor influences the value of (i,j), and $s_{kl;l}$ is the value of the pixel at the corresponding position (see Fig.2) where,

$s_{ij;1} = I_{i-1,j} + I_{i+1;j} \quad s_{ij;3} = I_{i-2,j} + I_{i+2;j} \quad s_{ij;5} = I_{i-1,j-1} + I_{i+1;j+1}$
$s_{ij;2} = I_{i,j-1} + I_{i;j+1} \quad s_{ij;4} = I_{i,j-2} + I_{i;j+2} \quad s_{ij;6} = I_{i-1,j+1} + I_{i+1;j-1}$

For an image segment of size $M$ and $N$ the GMRF parameters $\alpha$ and $\sigma$ are estimated using least square error estimation method, as follows:

$$\begin{pmatrix} \alpha_1 \\ \vdots \\ \alpha_n \end{pmatrix} = \left\{ \sum_{ij} \begin{bmatrix} s_{ij;1}s_{ij;1} & \cdots & s_{ij;1}s_{ij;n} \\ \vdots & \ddots & \vdots \\ s_{ij;n}s_{ij;1} & \cdots & s_{ij;n}s_{ij;n} \end{bmatrix} \right\}^{-1} \sum_{ij} I_{ij} \begin{pmatrix} s_{ij;1} \\ \vdots \\ s_{ij;n} \end{pmatrix} \quad (2)$$

$$\sigma^2 = \frac{\sum_{ij}\left[I_{ij} - \sum_{l=1}^{n}\alpha_l s_{ij;l}\right]}{(M-2)(N-2)} \quad (3)$$

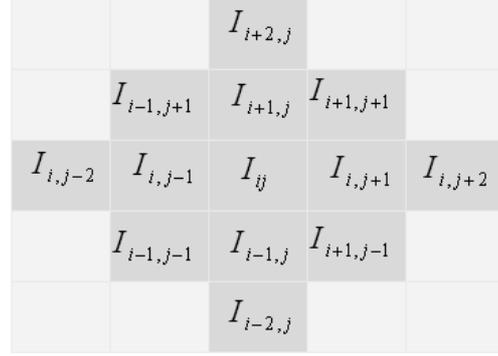

Fig. 2. Third order Markov neighbourhood for each sample image pixel ($I_{ij}$)

*Fractals*

Fractals are used to describe non-Euclidean structures that show self-similarity at different scales [4]. There are several fractal models used to estimate the fractal dimension; the fBm which is the mean absolute difference of pixel pairs as a function of scale as shown in (4) was adopted [5].

$$E(\Delta I) = K \Delta r^H \quad (4)$$

Where $\Delta I = |I(x_2,y_2) - I(x_1,y_1)|$ is the mean absolute difference of pixel pairs; $\Delta r = [(x_2 - x_1) + (y_2 - y_1)]^{1/2}$ is the pixel pair distances; $H$ is called the Hurst coefficient; and $K$ is a constant.
The fractal dimension (FD) can be then estimated by plotting both sides of (4) on a log-log scale and $H$ will represent the slope of the curve that is used to estimate the FD as: $FD = 3 - H$

By operating pixel by pixel, an FD image was generated for each sample image segment where each pixel has its own FD value. Then first order statistical features were derived, which are: mean variance, lacunarity (i.e., variance divided by mean), skewness and kurtosis.

*2.2.2 Statistical-based features methods*

*Co-occurrence matrices*

The Grey level co-occurrence matrix (GLCM) $P_g(i,j | \theta, d)$ represents the joint probability of certain sets of pixels having certain grey-level values. It calculates how many times a pixel with grey-level $i$ occurs jointly with another pixel having a grey value $j$. By varying the displacement vector $d$ between each pair of pixels many GLCMs with different directions can be generated. For each sample image segment and with distance set to one, four GLCMs

| Texture type | Texture feature extraction method ||||||||||
|---|---|---|---|---|---|---|---|---|---|---|
| | GLCM || GMRF || RLM || fBm || ACF ||
| | Training set | Testing set | Training set | Testing set | Training set | Testing set | Training set | Testing set | Training set | Testing set |
| D16 | 100% | 82.81% | 100% | 99.48% | 100% | 77.08% | 98.44% | 83.33% | 7.81% | 1.56% |
| D20 | 100% | 100% | 100% | 98.96% | 100% | 100% | 100% | 80.73% | 95.31% | 86.46% |
| D74 | 100% | 100% | 98.44% | 83.85% | 98.44% | 98.44% | 100% | 97.40% | 21.88% | 25.52% |
| D24 | 100% | 88.54% | 95.31% | 91.67% | 100% | 85.94% | 96.88% | 76.56% | 21.88% | 12.50% |
| D93 | 100% | 99.48% | 100% | 98.44% | 98.44% | 88.02% | 90.63% | 83.85% | 56.25% | 54.69% |
| D98 | 100% | 97.40% | 100% | 100% | 100% | 92.19% | 93.75% | 96.35% | 4.69% | 0.52% |
| D106 | 100% | 97.92% | 100% | 99.48% | 100% | 98.44% | 92.19% | 76.04% | 45.31% | 41.15% |
| D112 | 100% | 100% | 89.06% | 91.67% | 100% | 89.06% | 95.31% | 91.67% | 65.63% | 45.31% |
| accuracy | 100% | **95.77%** | 97.85% | **95.44%** | 99.61% | **91.15%** | 95.90% | **85.74%** | 39.84% | **33.46%** |

TableI: Overall accuracy of classification using each texture feature extraction method individually having directions (0°,45°,90° &135°) were generated. Having the GLCM normalized, we can then derived eight second order statistic features which are also known as haralick features [6] for each sample, which are: contrast, correlation, energy, entropy, homogeneity, dissimilarity, inverse difference momentum, maximum probability.

*Run-length matrices*

The grey level run-length matrix (RLM) $P_r(i, j | \theta)$ is defined as the numbers of runs with pixels of gray level $i$ and run length $j$ for a given direction $\theta$ [7]. RLMs was generated for each sample image segment having directions (0°,45°,90° &135°), then the following five statistical features were derived: short run emphasis, long run emphasis, gray level non-uniformity, run length non-uniformity and run percentage.

*Autocovariance function*

The Autocovariance function (ACF) is the autocorrelation function after subtracting the mean. It is a way to investigate non-randomness by looking for replication of certain patterns in an image. The ACF is defined as:

$$\rho(x,y) = \frac{\sum_{i=1}^{M-x}\sum_{j=1}^{N-y}(I(i,j)-\mu)(I(i+x,j+y)-\mu)}{(M-x)(N-y)} \quad (5)$$

Where $I(i, j)$ is the grey value of a $M \times N$ image, $\mu$ is the mean of the image before processing and $x, y$ are the amount of shifts.

After calculating the ACF for each sample image segment the peaks of the horizontal and vertical margins were fitted using least squares by an exponential function. Therefore, each sample is represented by four different parameters, which are the horizontal and vertical margins values referring to the ACF and exponential fittings.

### 2.3 Classification algorithm

The naïve Bayesian classifier (nBC) is a simple probabilistic classifier which assumes attributes are independent. Yet, it is a robust method with on average has a good classification accuracy performance, and even with possible presence of dependent attributes [8]. From Bayes' theorem,

$$P_i(C_i/X) = \frac{P(X/C_i)P(C_i)}{P(X)} \quad (6)$$

Given a data sample $X$ which represent the extracted texture features vector $(f_1, f_2, f_3 \ldots f_j)$ having a probability density function (PDF) $P(X/C_i)$, we tend to maximize the posterior probability $P(C_i/X)$ (i.e., assign sample $X$ to the class $C_i$ that yields the highest probability value).

Where $P(C_i/X)$ is the probability of assigning class $i$ given feature vector $X$; and $P(X/C_i)$ is the probability; $P(C_i)$ is the probability that class $i$ occurs in all data set; $P(X)$ is the probability of occurrence of feature vector $X$ in the data set.

$P(C_i)$ and $P(X)$ can be ignored since we assume that all are equally probable for all samples. This yields the maximum of $P(C_i/X)$ is equal to the maximum of $P(X/C_i)$ and can be estimated using maximum likelihood after assuming a Gaussian PDF [9] as follows:

$$P(X/C_i) = \frac{1}{(2\pi)^{n/2}|\Sigma_i|^{1/2}} \exp\left[-\tfrac{1}{2}(X-\mu_i)^T \sum_i^{-1}(X-\mu_i)\right]$$

(7)

| Combined methods | No. of features | Train set accuracy | Test set accuracy |
|---|---|---|---|
| GMRF & fBm | 12 | 99.80% | 97.01% |
| GMRF & RLM | 27 | 100% | 96.94% |
| GMRF & GLCM | 39 | 100% | 96.55% |
| RLM & fBm | 25 | 100% | 95.70% |
| GLCM & ACF | 36 | 100% | 95.05% |
| fBm & GLCM | 37 | 100% | 94.66% |
| GMRF & ACF | 11 | 97.66% | 92.84% |
| GLCM & RLM | 52 | 100% | 92.12% |
| RLM & ACF | 24 | 99.61% | 89.58% |
| fBm & ACF | 9 | 96.48% | 85.48% |

Table II
Overall accuracy of texture feature extraction methods combined in pairs

Where $\Sigma_i$ and $\mu_i$ are the covariance matrix and mean vector of feature vector $X$ of class $C_i$; $|\Sigma_i|$ and $\Sigma_i^{-1}$ are the determinant and inverse of the covariance matrix; and $(X - \mu_i)^T$ is the transpose of $(X - \mu_i)$.

## 3. Experiment Results and Discussion

Initially, each method was applied individually to each of the eight different texture images to show which performs better classification accuracy. The overall classification accuracies for each of the eight Brodatz images are shown in Table 1. The GLCM and GMRF achieved the highest classification rate with 95.77% and 95.44%, while RLM and fBm scored 91.15% and 85.74%. All methods achieved relative good accuracy except for the ACF.

The method that achieved the lowest misclassification in all of the eight texture images was GMRF. The least accuracy was in texture 3 with 83.85% and that was due to large structure of the image texture which was beyond the size of the used third order neighborhood box to capture.

Then the highest classification accuracy which was achieved by the GLCM method was set as the accuracy improvement criteria to compare the performance of the next part, where texture features from different methods are combined together to investigate if they may assist in increasing accuracy rate.

It was found that using the two model-based texture features (GMRF and fBm) together improved the overall accuracy up to 97.01%. Also, when combining the statistical-based RLM and GLCM with the GMRF texture features it increased the overall accuracy to 96.94% and 96.55%; respectively. RLM with fBm gave nearly the same accuracy using GLCM alone, while the rest of combinations scored less than the predefined accuracy improvement criteria. Table 2 summarizes all classification accuracies and number of used texture features for all possible paired combinations.

It was also noticed that the GMRF texture features appear in all paired combinations which improved the overall accuracy, and with only 12 features, the GMRF and fBm combination overcame the accuracy achieved when using the GLCM individually which needs 32 features. Combining more than two methods with each other did not improve much the accuracy (e.g. GMRF with fBm and RLM achieved the highest with 97.07% classification accuracy) and will increase the time for computation as well. Accuracy could be further improved if a feature selection method was used to remove possibly highly correlated features; this needs to be further investigated.

## 4. Conclusion

As texture feature extraction methods tend to capture different image texture characteristics, using different combinations could assist in improving the classifier accuracy. Using a nBC, it was shown that combined model-based texture feature extraction methods (GMRF with fBm) proved to be better in classifying as compared to statistical methods. The model-based combined features improved the overall classification accuracy above the highest achieved using each of five different methods individually. Moreover, using GMRF features with statistical methods (RLM and GLCM) improved the overall accuracy as well.

## 5. References

[1] M. Tuceryan and A. Jain, *The Handbook of pattern Recognition and Computer Vision*, 2 ed: World Scientific Publishing Co., 1998.
[2] P. Brodatz, *A Photographic Album for Artists and Designers*. New York: Dover, 1996.
[3] M. Petrou and P. Gacia Sevilla, *Image processing: Dealing with texture*: Wiley, 2006.
[4] B. B. Mandelbrot, *Fractal Geometry of Nature*. San Francisco, CA: Freeman, 1982.
[5] C. C. Chen, J. S. Daponte, and M. D. Fox, "Fractal Feature Analysis and Classification in Medical Imaging," *IEEE Transactions on Medical Imaging*, vol. 8, pp. 133-142, 1989.
[6] R. M. Haralick, Shanmuga.K, and I. Dinstein, "Textural Features for Image Classification," *IEEE Transactions on Systems, Man and Cybernetics*, vol. SMC3, pp. 610-621, 1973.
[7] M. M. Galloway, "Texture analysis using gray level run lengths," *Computer Graphics Image Processing*, vol. 4, pp. 172-179, 1975.
[8] F. Demichelis, P. Magni, P. Piergiorgi, M. A. Rubin, and R. Bellazzi, "A hierarchical Naive Bayes Model for handling sample heterogeneity in classification problems: an application to tissue microarrays," *Bmc Bioinformatics*, vol. 7, 2006.
[9] R. C. Gonzales and R. E. Woods, *Digital Image Processing*, 2 ed: Prentice Hall, 2002.